\documentclass[10pt,twocolumn,letterpaper]{article}

\usepackage[pagenumbers]{cvpr}     % Anonymous submission

\definecolor{cvprblue}{rgb}{0.21,0.49,0.74}
\usepackage[pagebackref,breaklinks,colorlinks,allcolors=cvprblue]{hyperref}

\def\paperID{324}
\def\confName{CV4Animals}
\def\confYear{2026}

\title{Generating Synthetic Wildlife Health Data from Camera Trap Imagery: A Pipeline for Alopecia and Body Condition Training Data}

\author{David Brundage PhD\\
University of Wisconsin - Madison, School of Veterinary Medicine\\
\tt\small Brundage2@wisc.edu}

\begin{document}

\maketitle

\begin{abstract}
No publicly available, ML-ready datasets exist for wildlife health conditions in camera trap imagery, creating a fundamental barrier to automated health screening. We present a pipeline for generating synthetic training images depicting alopecia and body condition deterioration in wildlife from real camera trap photographs. Our pipeline constructs a curated base image set from iWildCam using MegaDetector-derived bounding boxes and center-frame-weighted stratified sampling across 8 North American species. A generative phenotype editing system produces controlled-severity variants depicting hair loss consistent with mange and emaciation. An adaptive scene-drift quality control system uses a sham pre-filter and decoupled mask-then-score approach with complementary day/night metrics to reject images where the generative model altered the original scene. We frame the pipeline explicitly as a screening data source. From 201 base images across 4 species, we generate 553 QC-passing synthetic variants with an overall pass rate of 83\%. A sim-to-real transfer experiment---training exclusively on synthetic data and testing on real camera trap images of suspected health conditions---achieves 0.85 AUROC, demonstrating that the synthetic data captures visual features sufficient for screening.
\end{abstract}

\section{Introduction}
\label{sec:intro}

Camera trap networks generate tens of millions of images per year~\cite{Steenweg2017}, and computer vision has enabled species identification~\cite{Norouzzadeh2017, Willi2018, Beery2018}, counting, and behavioral classification at population scale~\cite{Tuia2022}. Yet automated health screening remains absent. A biologist reviewing camera trap footage may notice that a coyote exhibits patchy hair loss suggestive of mange, or that a deer appears emaciated---but these observations are made opportunistically, are not systematically recorded, and cannot scale.

The barrier is data, not architecture. Prior work on camera trap health surveillance has been primarily manual: Oleaga~\etal~\cite{Oleaga2011} pioneered visual mange detection in wolves and foxes; subsequent studies labeled camera trap images for mange-compatible lesions in coyotes~\cite{Murray2021}, red foxes~\cite{CarricondoSanchez2017}, and other species~\cite{BarrosoPalencia2025}---but none released ML-ready image datasets. Ringwaldt~\etal~\cite{Ringwaldt2026} developed an automated system, training a CNN on $\sim$7,800 manually labeled crops of brushtail possums, but no known labeled data exists for North American species. On the synthetic side, Beery~\etal~\cite{Beery2020}, Lanzini and Beery~\cite{Lanzini2021}, and DisCL~\cite{Liang2024} have demonstrated that synthetic camera trap images improve generalization for species classification---but none generate images depicting health conditions. At this time, no known publicly available benchmark exists for training or evaluating automated wildlife health screening.

Visual assessment---whether by a human biologist or an AI system---cannot provide a clinical diagnosis from camera trap imagery alone. Alopecia in wildlife has numerous causes including sarcoptic and demodectic mange, lice, seasonal shedding, mechanical hair loss, and idiopathic follicular inactivity~\cite{Pence2002, Niedringhaus2019, WDFW2025}. Our pipeline generates training data labeled as ``hair loss consistent with mange'' rather than diagnosed mange, and the intended model output is a suspect flag for expert review. Lesion distribution varies by mange mite species and host taxon, with the face and periocular region being the typical initial site for sarcoptic mange in canids; confirmed mange in white-tailed deer is rare in some regions, where observed hair loss is more commonly attributable to shedding or follicular inactivity.

Our contributions are: (1)~the first pipeline for generating synthetic wildlife health training data from camera trap imagery, with adaptive quality control; (2)~a phenotype design that addresses species-specific lesion distribution and explicitly frames outputs as screening data; and (3)~a sim-to-real transfer experiment demonstrating that a classifier trained exclusively on synthetic images can detect suspected health conditions in real camera trap photographs.
\section{Pipeline}
\label{sec:pipeline}

The pipeline has three stages: each base image is filtered from iWildCam and localized with MegaDetector, edited into up to four phenotype variants by a generative model, and quality-controlled for scene preservation.Synthetic dataset will be released at \url{https://huggingface.co/datasets/BrundageLab/synthetic_wildlife_health}

\subsection{Base Image Curation}
\label{sec:curation}

\textbf{Species scope.} The pipeline targets 8 North American species selected for their relevance to wildlife health monitoring and representation in iWildCam 2022~\cite{Beery2022}: white-tailed deer (\textit{Odocoileus virginianus}), mule deer (\textit{O.~hemionus}), elk (\textit{Cervus canadensis}), gray wolf (\textit{Canis lupus}), red fox (\textit{Vulpes vulpes}), gray fox (\textit{Urocyon cinereoargenteus}), raccoon (\textit{Procyon lotor}), and a broad cervid category.

\textbf{Image selection.} MegaDetector v4~\cite{Beery2019} provides animal bounding boxes (highest-confidence animal detection per image). We construct a balanced 1,000-image base set via weighted sampling that prioritizes center-frame animal placement (weight 0.6 for center, 0.3 for mid-frame, 0.1 for edge/corner positions based on normalized bounding box centroid) while maintaining proportional species representation and temporal diversity across seasons and day/night conditions. For each generation run, a stratified subsample is drawn preserving the species $\times$ day/night $\times$ season distribution.

\subsection{Phenotype Editing}
\label{sec:editing}

\begin{figure*}[t]
\centering
\includegraphics[width=\textwidth]{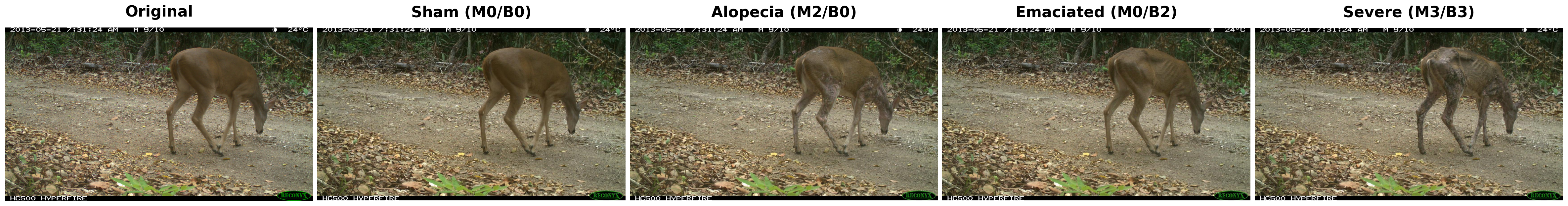}

\caption{Synthetic phenotype variants for white-tailed deer (\textit{Odocoileus virginianus}). From left: original camera trap image and four QC-passing edits at increasing severity. The sham (M0/B0) serves as negative control. Alopecia (M2/B0) depicts patchy hair loss; emaciated (M0/B2) depicts visible skeletal landmarks; severe (M3/B3) combines extensive hair loss with emaciation. }
\label{fig:phenotype}
\end{figure*}

Each base image yields up to four synthetic variants via Gemini 3.1 Flash Image~\cite{Gemini2025, Brooks2023}, spanning two visual dimensions. (Fig.~\ref{fig:phenotype}).

\textbf{Alopecia severity.} Three levels: \textbf{M0} (healthy intact coat), \textbf{M2} (moderate patchy hair loss with scaling and crusting), and \textbf{M3} (severe extensive hair loss with hyperkeratotic plaques and lichenification). We label these as ``hair loss consistent with mange''.

\textbf{Body condition score.} Three levels: \textbf{B0} (healthy musculature), \textbf{B2} (visible ribs, noticeable muscle wasting), and \textbf{B3} (severe emaciation with pronounced skeletal landmarks).

\textbf{Variant design.} The four variants per base image are: sham (M0/B0, negative control), alopecia-only (M2/B0), emaciated-only (M0/B2), and severe-both (M3/B3).

\textbf{Lesion distribution.} For canids, the edit prompt specifies the typical sarcoptic mange progression: initial involvement at the face (particularly periocular regions), ear margins, and elbows, extending to flanks and legs at moderate severity, and generalized at severe stages~\cite{Pence2002, Niedringhaus2019}. For hoofstock, lesion distribution accounts for the anatomy of self-grooming---areas beyond the shoulder blades along the spine are typically spared because they are inaccessible to scratching with teeth or hind hooves.

\textbf{Edit prompt design.} Each prompt provides the generative model with the MegaDetector-derived normalized bounding box coordinates, species identity, severity labels with descriptive text, and explicit instructions to edit only the animal within the bounding box while preserving the background, lighting, noise pattern, timestamps, and camera-trap artifacts pixel-identical. This spatial constraint---anchoring edits to the detected animal region---is critical for the QC system (Sec.~\ref{sec:qc}), as it establishes a clear expected boundary between edited and preserved regions.

\subsection{Scene-Drift Quality Control}
\label{sec:qc}

\begin{figure*}[t]
\centering
\includegraphics[width=\textwidth]{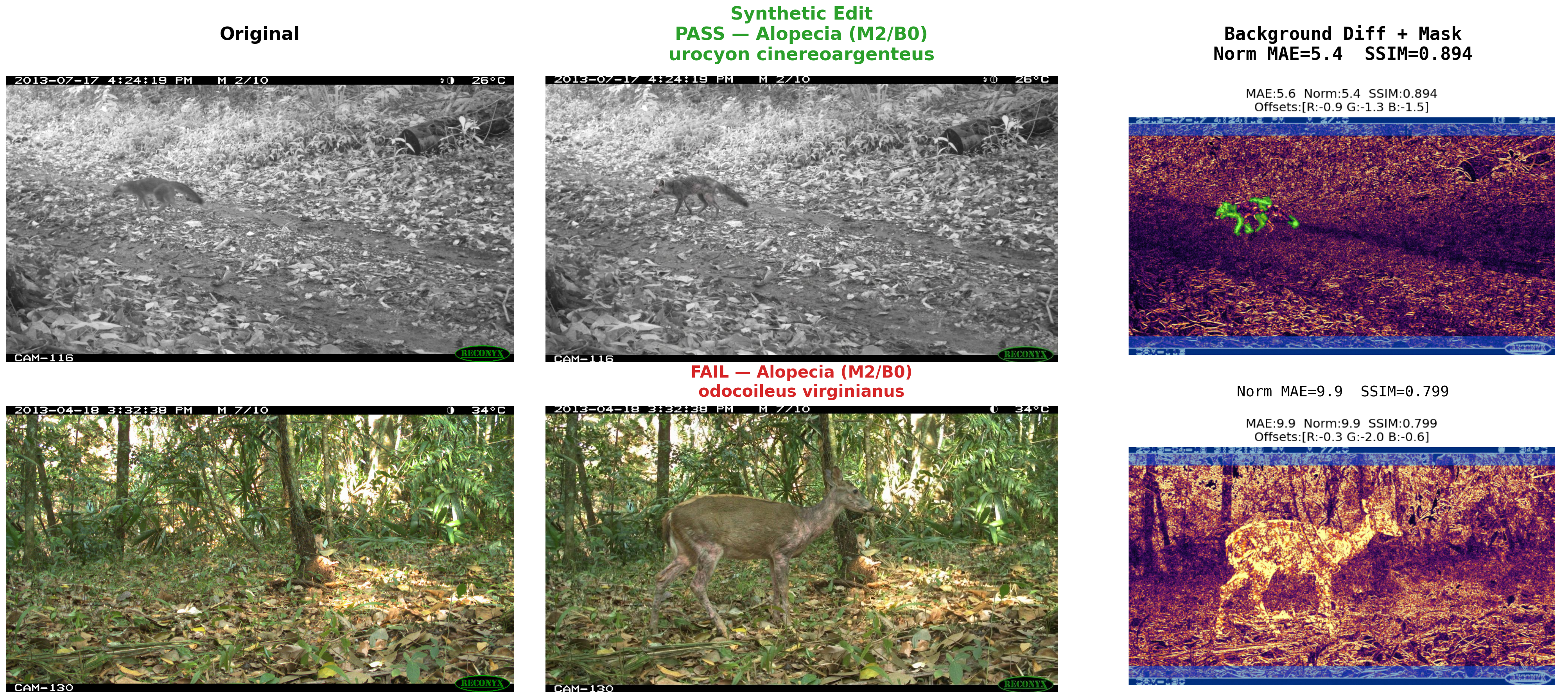}
\caption{Both rows show the same alopecia edit (M2/B0). (A) Scene preserved: the change mask (green) is confined to the animal region; metrics pass both thresholds. (B) The generative model hallucinated and rendered a completely different animal outside the bounding box region.}
\label{fig:qc}
\end{figure*}

The QC system determines whether the generative model preserved the scene (Fig.~\ref{fig:qc}). It uses a decoupled design: the animal mask is built from raw pixels, then quality is scored on blurred pixels outside the mask.

\textbf{Sham pre-filter.} Experiments revealed that approximately 16\% of base images produce scenes that the generative model fundamentally re-renders. We generate the sham (M0/B0) variant first; if it fails QC, all remaining variants for that base are skipped. This saves generation budget by detecting irrecoverable scenes early.

\textbf{Mask construction.} Raw pixel differences between original and synthetic images are thresholded, and connected component labeling identifies changed regions. Small blobs are discarded as compression noise, and remaining blobs are retained only if they overlap the MegaDetector bounding box, anchoring the mask to the known animal location. The mask is conservatively dilated; images where it exceeds 70\% of the image area are rejected as global re-renders. Top and bottom edge margins are excluded from scoring to avoid camera-trap timestamp overlays.

\textbf{Scene scoring.} Three complementary metrics are computed on Gaussian-blurred (radius~2) scene pixels: raw mean absolute error (MAE), color-normalized MAE (per-channel DC offset removed before scoring, correcting for the generative model's systematic warm/cool color shifts), and structural similarity index (SSIM, window size~7)~\cite{Wang2004}. Adaptive thresholds account for different day/night characteristics. \textit{Daytime:} an OR-gate passes images if normalized MAE $\leq$ 7.0 OR SSIM $\geq$ 0.85, exploiting complementary failure modes---normalized MAE handles global color shifts that SSIM tolerates, while SSIM handles localized texture re-rendering that inflates MAE. \textit{Nighttime/IR:} raw MAE $\leq$ 5.0 (stricter, as backgrounds change minimally under infrared illumination).

\section{Results}
\label{sec:results}

\begin{table}[t]
\centering
\small
\begin{tabular}{@{}lr@{}}
\toprule
Metric & Value \\ \midrule
Species represented & 4 \\
Base images processed & 201 \\
Total variants generated & 666 \\
QC-passing variants & 553 (83\%) \\
Sham pre-filter rejections & 31/191 (16\%) \\
Variants skipped via sham & 93 \\
Daytime pass rate & 85\% \\
Nighttime pass rate & 81\%  \\ \bottomrule
\end{tabular}
\caption{Synthetic data generation summary. The sham pre-filter saved 93 variant-level API calls by detecting irrecoverable scenes early. QC pass rates are consistent across day and night conditions.}
\label{tab:stats}
\end{table}

Table~\ref{tab:stats} summarizes the pipeline output. From a stratified subsample of 201 base images across 4 species (gray fox, gray wolf, white-tailed deer, raccoon), the pipeline generated 666 synthetic variants, of which 553 (83\%) passed scene-drift QC. The sham pre-filter rejected 16\% of base images, saving 93 variant-level API calls that would have produced unusable outputs.

\begin{table}[t]
\centering
\footnotesize
\begin{tabular}{@{}lccccc@{}}
\toprule
Variant & Pass\% & Norm & SSIM & Both & $\bar{\text{SSIM}}$ \\ \midrule
Sham (M0/B0) & 78\% & 0 & 4 & 74 & 0.900 \\
Alopecia (M2/B0) & 90\% & 0 & 10 & 60 & 0.889 \\
Emaciated (M0/B2) & 92\% & 0 & 10 & 63 & 0.889 \\
Severe (M3/B3) & 82\% & 0 & 11 & 55 & 0.886 \\ \bottomrule
\end{tabular}
\caption{QC pass rates by variant type (daytime images, $N{=}337$). Columns show how many images passed via each OR-gate path: normalized MAE only, SSIM only, or both. Nearly all daytime passes satisfy both criteria simultaneously. Mean SSIM decreases with edit severity, confirming the QC metrics are sensitive to edit magnitude.}
\label{tab:qc}
\end{table}

Table~\ref{tab:qc} shows QC pass rates broken down by variant type for daytime images. The severe-both variant (M3/B3) has the lowest pass rate (82\%) and lowest mean SSIM (0.886), consistent with the expectation that larger edits produce more scene drift. Nearly all daytime passes satisfy both the normalized MAE and SSIM criteria simultaneously, indicating that the OR-gate rarely rescues images that would fail a single-metric threshold. Species-level pass rates range from 100\% (gray wolf, single base image) to 50--67\% for raccoon severe variants, while the dominant species (gray fox, $N{=}136$ base images) achieves 78--88\% across variants.
\subsection{Sim-to-Real Transfer}

\begin{figure}[t]
\centering
\includegraphics[width=\columnwidth]{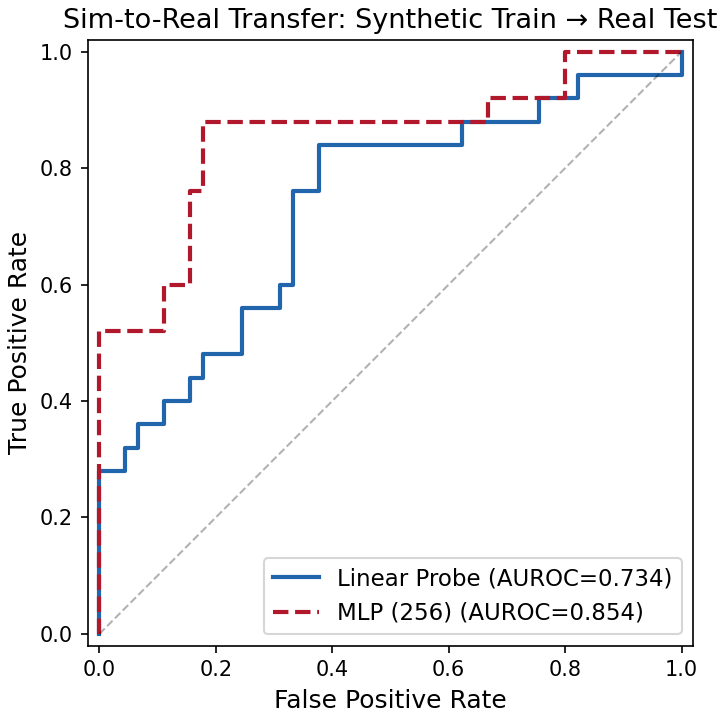}

\caption{Receiver operating characteristic for sim-to-real transfer. Both heads trained on synthetic data only and evaluated on real camera trap images. The MLP (AUROC = 0.854) substantially outperforms the linear probe (0.734)}
\label{fig:roc}
\end{figure}

To test whether the synthetic data captures visual features relevant to real health conditions, we conduct a sim-to-real transfer experiment: train \textit{exclusively} on synthetic images, test \textit{exclusively} on real images (Fig.~\ref{fig:roc}). We extract features from a frozen DINOv2 ViT-B/14~\cite{Oquab2023} backbone (768-dimensional CLS token) and train two classification heads: a linear probe (logistic regression) and a single-layer MLP (256 hidden units), both with balanced class weights. No real images are used during training.

The training set consists of 553 QC-passing synthetic variants: 160 sham images (healthy, label~0) and 393 phenotype variants (suspect, label~1). The test set consists of 70 real camera trap images: 45 healthy controls sampled from iWildCam images excluded from the pipeline's base image pool (to prevent the model from exploiting shared background or scene-level cues), and 25 images depicting suspected health conditions sourced from trail camera and camera trap community forums, all from the same target species as training. Both healthy and suspect test images are camera trap photographs, controlling for sensor modality. Suspect labels reflect community observer reports, not clinical confirmation, consistent with the screening framing of this work. The suspect class contains deer (14), wolf (5), fox (3), raccoon (2), and bear (1); the healthy class contains the same core species (28 white-tailed deer, 10 gray wolf, 6 gray fox, 1 raccoon), ensuring that species identity alone cannot discriminate the classes.

 The MLP achieves 0.854 AUROC and 0.811 balanced accuracy, demonstrating that synthetic data alone is sufficient to detect suspected health conditions in real camera trap images the model has never seen. The linear probe achieves 0.734 AUROC, indicating that the visual signal distinguishing healthy from suspect animals is partially but not fully linearly separable in DINOv2 feature space---health assessment benefits from composing multiple visual cues (coat texture, body shape, skeletal prominence) that a single hidden layer can capture. Five-fold cross-validation on the synthetic training set yields 0.830 $\pm$ 0.032 AUROC, confirming the synthetic images are internally consistent and learnable.

\section{Discussion}
\label{sec:discussion}

\textbf{Limitations.} (1)~\textit{Diagnostic ambiguity}---visual assessment cannot distinguish mange from lice, shedding, or idiopathic hair loss, which is fundamental to the modality, not unique to our approach. (2)~\textit{Species-condition mismatch}---confirmed mange in white-tailed deer is rare in some regions, so the model may flag seasonal or mechanical hair loss, requiring per-region false-positive calibration. (3)~\textit{Posture confounds}---editing onto healthy base images cannot capture illness-related posture changes that biologists use as supporting evidence; future versions could incorporate posture modification using pose-conditioned generative models, though this would require validation by wildlife pathologists. (4)~\textit{Transfer evaluation}---the real test set is small ($N{=}70$) and suspect labels are community-reported, not clinically confirmed; the 0.85 AUROC should be interpreted as evidence of feasibility rather than a production-ready performance estimate. 

\textbf{Future work.} Preliminary deployment on 22,958 Snapshot Wisconsin \cite{SnapshotWisconsin2024} images flagged ${\sim}$5\% for expert review. Planned extensions include pathologist validation of synthetic image realism, a larger evaluation with clinical cases.

{
    \small
    \bibliographystyle{ieeenat_fullname}
    \bibliography{main}
}

\end{document}